\documentclass[conference]{IEEEtran}
\IEEEoverridecommandlockouts
\usepackage{cite}
\usepackage{amsmath,amssymb,amsfonts}
\usepackage{algorithmic}
\usepackage{graphicx}
\usepackage{textcomp}
\usepackage{xcolor}
\def\BibTeX{{\rm B\kern-.05em{\sc i\kern-.025em b}\kern-.08em
T\kern-.1667em\lower.7ex\hbox{E}\kern-.125emX}}
\usepackage[breaklinks]{hyperref}

\newcommand{\linebreakand}{%
\end{@IEEEauthorhalign}
\hfill\mbox{}\par
\mbox{}\hfill\begin{@IEEEauthorhalign}
}

\renewcommand\IEEEkeywordsname{Keywords}
\begin{document}

\title{SwarmHive: Heterogeneous Swarm of Drones for Robust Autonomous Landing on Moving Robot\\
\thanks{The reported study was funded by RFBR and CNRS according to the research project No. 21-58-15006.}
}

\author{\IEEEauthorblockN{Ayush Gupta} 
\IEEEauthorblockA{
\textit{Skoltech}\\
Moscow, Russia \\
ayush.gupta@skoltech.ru}
\and
\IEEEauthorblockN{Ahmed Baza}
\IEEEauthorblockA{
\textit{Skoltech}\\
Moscow, Russia \\
ahmed.baza@skoltech.ru}
\and

\IEEEauthorblockN{Ekaterina Dorzhieva}
\IEEEauthorblockA{
\textit{Skoltech}\\
Moscow, Russia \\
ekaterina.dorzhieva@skoltech.ru}
\and
\IEEEauthorblockN{Mert Alper}
\IEEEauthorblockA{
\textit{Skoltech}\\
Moscow, Russia \\
mert.alper@skoltech.ru}
\and 

\IEEEauthorblockN{Mariia Makarova}
\IEEEauthorblockA{
\textit{Skoltech}\\
Moscow, Russia \\
mariia.makarova@skoltech.ru}
\and
\IEEEauthorblockN{Stepan Perminov}
\IEEEauthorblockA{
\textit{Skoltech}\\
Moscow, Russia \\
stepan.perminov@skoltech.ru}
\and
\IEEEauthorblockN{Aleksey Fedoseev}
\IEEEauthorblockA{
\textit{Skoltech}\\
Moscow, Russia \\
aleksey.fedoseev@skoltech.ru}
\and
\IEEEauthorblockN{Dzmitry Tsetserukou}
\IEEEauthorblockA{
\textit{Skoltech}\\
Moscow, Russia \\
d.tsetserukou@skoltech.ru}

}

\maketitle

\begin{abstract}
The paper focuses on a heterogeneous swarm of drones to achieve a dynamic landing of formation on a moving robot. This challenging task was not yet achieved by scientists. The key technology is that instead of facilitating each agent of the swarm of drones with computer vision that considerably increases the payload and shortens the flight time, we propose to install only one camera on the leader drone. The follower drones receive the commands from the leader UAV and maintain a collision-free trajectory with the artificial potential field.    

The experimental results revealed a high accuracy of the swarm landing on a static mobile platform (RMSE of 4.48 cm). RMSE of swarm landing on the mobile platform moving with the maximum velocities of 1.0 m/s and 1.5 m/s equals 8.76 cm and 8.98 cm, respectively.
The proposed SwarmHive technology will allow the time-saving landing of the swarm for further drone recharging. This will make it possible to achieve self-sustainable operation of a multi-agent robotic system for such scenarios as rescue operations, inspection and maintenance, autonomous warehouse inventory, cargo delivery, and etc. 
\end{abstract}


\section{Introduction}
\begin{figure}[!h]
 \centering
 \includegraphics[width=1\linewidth]{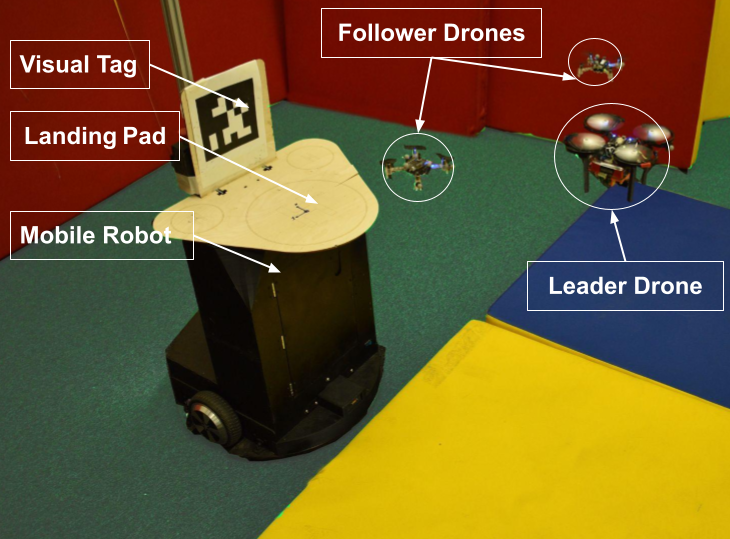} 
 \caption{Heterogeneous swarm of drones SwarmHive accomplishes landing on the surface of mobile robot.}
 \label{fig:realsystem}
\end{figure}

The recent development in Unmanned Aerial Vehicle (UAV) control systems has led to their extensive application in search and rescue, sensor networks in the Internet of Things (IoT), remote sensing surveillance, inspection, and etc \cite{Eaton_2016, Kangunde_2021}.
The autonomous delivery solutions with heterogeneous mobile robot teams are extensively investigated by delivery companies such as Amazon Technologies Inc., which presented concepts of multi-robotic ground delivery \cite{Scott_2020} and fulfillment center with UAV delivery \cite{Curlander_2017}. Drone swarms may perform aforementioned actions autonomously, significantly increasing the variety of cargo size and saving time in comparison with single drone. 
However, the limited flight time of drones and the risk of their crashing during landing present a high entry threshold for the on-site UAV systems. The swarm delivery systems should, therefore, support the safe behavior of its agents in uncertain conditions, including taking off and landing on uneven surfaces, sustaining the communication between agents, and docking on both static and moving vehicles.

Several scenarios of UAV landing with high tolerance to the surface were introduced in the prior research. An origami-inspired cargo drone with a foldable cage was suggested by Kornatowski et al. \cite{Kornatowski_2017} for safe landing and interaction with humans. 
Sarkisov et al. \cite{Sarkisov_2018} proposed the landing gear with robotic legs, allowing drones to adapt to the uneven surface. The embedded torque sensors help this system to maintain stable landing leveraging the data on the contact forces with the ground. 

However, the mentioned approaches still require the assistance of a human operator during the deployment and docking of the swarm. An effective UAV parcel handover system with a high-speed vision system for the supply station control has been introduced by Tanaka et al. \cite{Tanaka_2019}. This research also suggests a non-stop UAV delivery concept based on a high-speed visual control with a 6-axis robot. The concept of landing the swarm of nano-copters on the human arms for prompt and safe deployment was proposed by Tsykunov et al. \cite{Tsykunov_2020}. Midair docking of the swarm with a robot arm equipped with a force-sensitive soft gripper is explored by Fedoseev et al. \cite{Fedoseev_2021}. The downside of the aforementioned methods is that they require either a static landing platform or a moving platform that actively seeks to catch the drone. In the following section, we describe the state-of-the-art systems for landing on the dynamic platform that moves independently.

\section{Related Works}

The complexity of the real-time landing on a mobile platform adds additional restrictions to the system. First of all, it is important to maintain the connection between the swarm agents and to prevent their loss of formation. Secondly, such systems require higher precision of the CV-based localization due to the limited size of the landing pad that should be shared by the swarm without any internal collision of its units. The latter problem has been extensively explored in the case of single drone deployment and landing in various environments. For example, Muskardin et al. \cite{Muskardin_2016} developed a docking system with a fixed-wing Penguin BE UAV capable of landing on the roof of a moving car. Baca et al. \cite{Baca_2017} developed a UAV with fast onboard detection of moving landing platform. Landing a multi-rotor UAV on a mobile robot with visual estimation of the target has been investigated by Araar et al. \cite{Araar_2017}. Scenarios for autonomous landing on an unmanned ground vehicle have been proposed by Wang et al. \cite{Wang_2018} and Shao et al. \cite{Shao_2019}.

The Computer Vision (CV) system plays an important role in a majority of outdoor landing systems since real-time target tracking is required. Color-based detection algorithms \cite{Respall_2019} or visually distinctive tagging \cite{Falanga_2017} have been applied to recognize a moving platform. Several works merges GPS and CV approaches, e.g., Feng et al. \cite{Feng_2018} observed the landing platform by the camera at a close distance and relied on GPS data otherwise. To increase the range and accuracy of the target vehicle recognition, Xing et al. \cite{Xing_2019} proposed a composite landmark. Robust autonomous drone take-off and landing on a moving platform were developed by Palafox et al. \cite{Palafox_2019}. Alijani et al. \cite{Alijani_2020} implemented a mathematical approach in the X-Y plane based on the inclination angle and state of the UAV for a safe landing on a moving vehicle. Kalinov et al. \cite{Kalinov_2019} introduced a high-precision visual recognition of infrared (IR) markers on a mobile robot and impedance-based control to achieve soft landing of a drone on a static or moving robot.

The influence of external disturbances on the landing platform and the UAV has been partially investigated. For example, Xuan-Mung et al. \cite{Xuan-Mung_2020} have presented a robust altitude control algorithm, a landing target state estimator, and an autonomous precision landing planner to cope with disturbances. Landing in turbulent wind conditions was presented by Paris et al. \cite{Paris_2020}.

While previously developed systems have explored single drone landing, the case of multi-agent landing presents additional challenges. The concept of swarm topology for docking was therefore proposed by Tahir et al. \cite{Tahir_2020}. Leon-Blanko et al. \cite{LeonBlanko_2022} investigated the issue of a multi-drone team logistics, where a team of UAVs visits a set of points with a delivery truck acting as a docking station. However, these systems are focused on the hierarchy and communication stability in the swarm. Meanwhile, the problem of multi-agent CV-based behavior while landing on a limited area surface of a moving platform was not yet explored and experimentally evaluated for both low-scale and large-scale swarms.


In this paper, we propose a SwarmHive technology (see Fig. \ref{fig:realsystem}), i.e. a heterogeneous swarm of drones capable of landing on a moving ground robot with a small landing area. Additionally, we evaluate the CV-based approach both on a static and dynamic platform. In the last section, possible improvements to the system are discussed.  
\section{SwarmHive System Overview}

\begin{figure}[!h]
 \centering
 \includegraphics[width=1\linewidth]{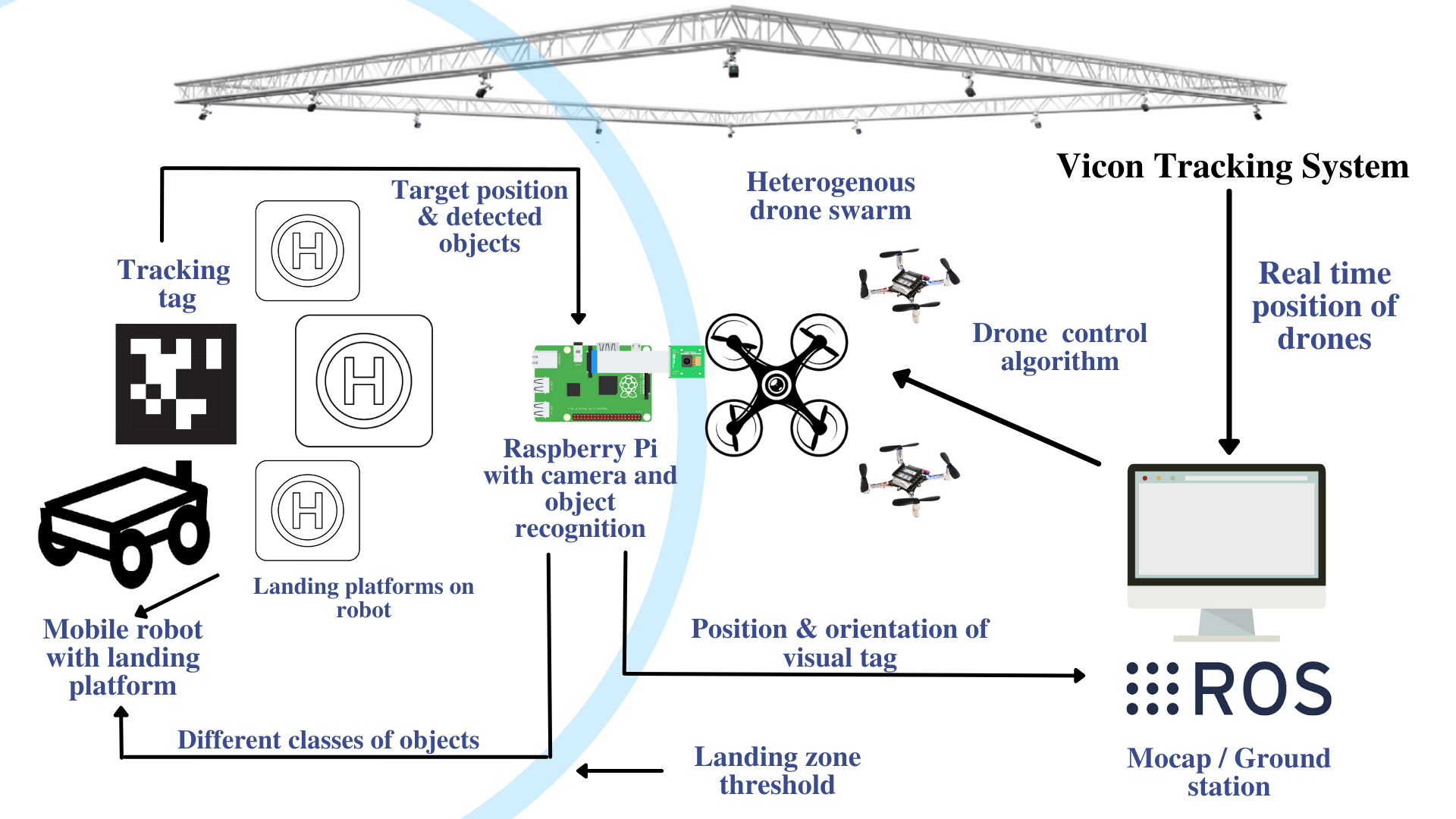} 
 \caption{Layout of the SwarmHive system.}
 \label{fig:systemarc}
\end{figure}

SwarmHive system consists of a mobile robot (mobile landing platform) with a visual tag marking the landing platform position and a swarm of drones that are targeting to land on the mobile robot (see Fig. \ref{fig:systemarc}). 

The swarm of drones is made up of two follower drones and a leader drone equipped with onboard processor (Raspberry Pi 4 with 4 GB RAM) and camera (8MP Pi camera Ver. 2.1) that are used to locate the landing platform position. VICON Vantage V5 motion capture system is used for localizing the micro drones and sending their updated position to the ground station. The ground station PC receives the updated positions of all drones from mocap system and the landing target position from the leader drone through ROS topics and then calculates the swarm target trajectory using continuous feedback-based planning.

Artificial Potential Field (APF) approach (discussed in \ref{section:APF}) ensures the avoidance of collision between the swarm agents. The resulting setpoints are sent to Crazyflie onboard controller using the Crazyswarm ROS package. The leader drone tracks the landing platform following the robot from a fixed altitude until the swarm is within the landing threshold. All agents start descending (in formation) towards the platform shutting the motors down when landing is achieved. The proposed SwarmHive system is capable of landing on a moving platform while avoiding inter-agent collision using visual tracking of the landing zone and APF.



\subsection{Vision-based Tracking Measurements}
\begin{figure}[!h]
 \centering
 \includegraphics[width=0.6\linewidth]{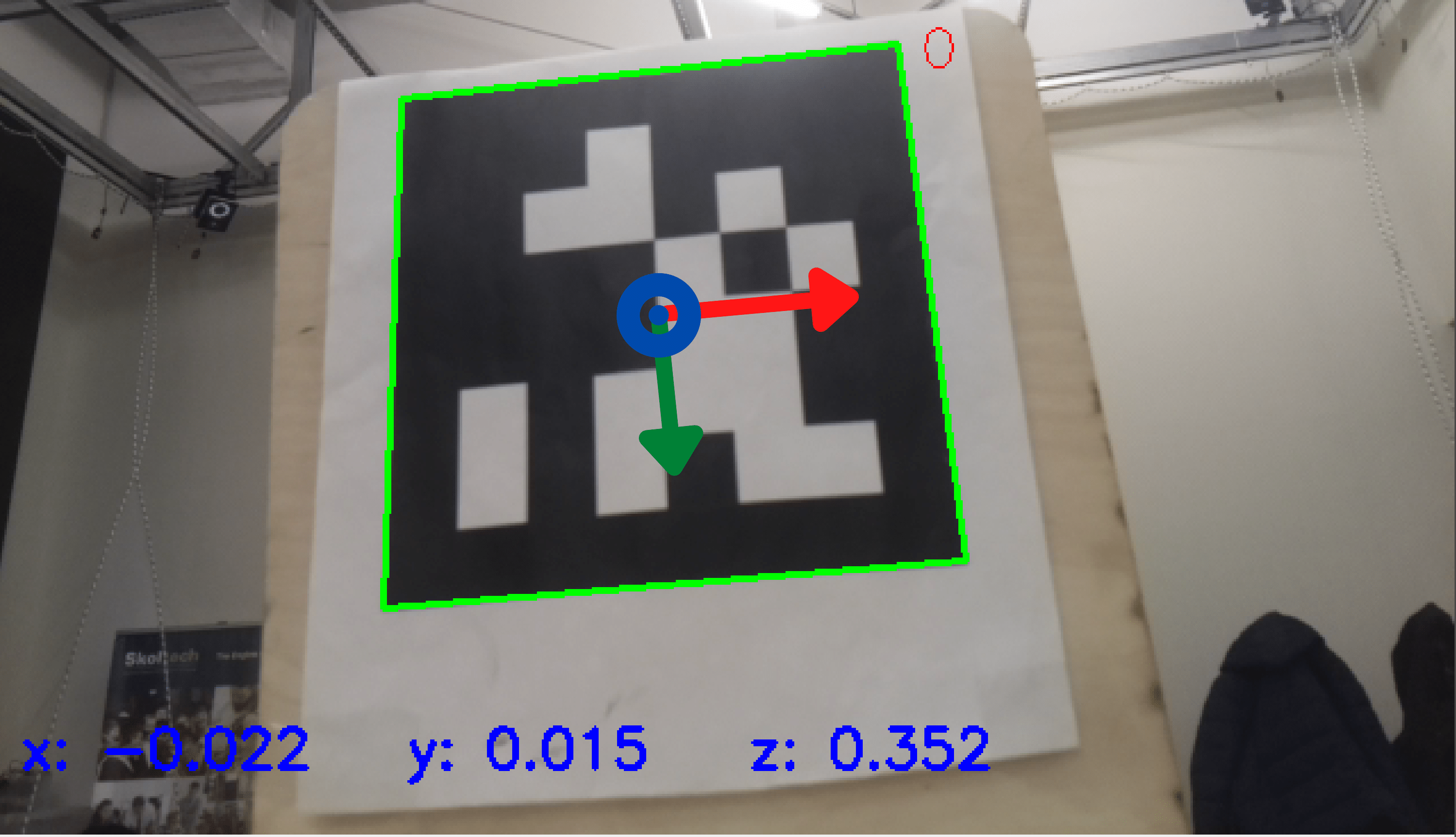} 
 \caption{Tracking of Apriltag marker with frame origin located in the center of the marker and Z-axis oriented towards the drone.}
 \label{fig:apriltag}
\end{figure}

For the detection of the target landing site on the moving platform, we used the most recent visual fiducial system AprilTag3 with OpenCV \cite{apriltag} and ROS as a bridge to pass the coordinates of fiducial markers (see Fig. \ref{fig:apriltag}).

With the help of this system, we estimate the position of possible tags on the image using a graph-based image segmentation algorithm and obtain a visual estimation of possible tags more accurately than GPS systems. In addition, we tested the algorithm with tags of different sizes and opted for the size of 16.6 x 16.6 cm based on the robust tag detection at wide range of distances. The tag can be detected by a leader drone at a maximum distance of approximately 3.5 to 4.0 m and a minimum distance of 30 cm. 

By tracking several feature points between two consecutive frames of a video, we can robustly estimate their motion and then compensate for it. The tracking algorithm works better with textured regions and corners, assuming that pixel intensities of an object are the same with every frame of the video, and we can track similar motion in neighboring pixels. To find the difference in intensity for a displacement in all directions between the original and moved window, Lucas-Kanade Optical Flow method was applied \cite{Lucas_1981}.





After finding features and applying optical flow, a rigid transform of the feature maps from the previous frame to the current frame of the video is calculated by using two sets of points. 

\subsection{Target Following with Artificial Potential Field}
\label{section:APF}
To avoid internal swarm collisions while landing on the small landing pad, we applied continuous feedback-based planning implementing an APF policy as our navigation function to achieve a collision-free trajectory from the starting position of drones to the position of rover \cite{Li_2020}. APF models a swarm agent as a point affected by the forces of two fields: the repulsion field and attraction field. The artificial repulsive field is generated by the obstacles (in this case the rest of swarm agents), while the artificial attraction field is the result of the target point. By summing two potentials and integrating the equations of motion of the modeled point we obtain the collision-free path for each agent from its current position to the target position. The APF path is recalculated along the way to guarantee collision-fee trajectory. The equations defining APF are shown in the following equations:

\vspace{-0.4em}
\begin{equation}
 \label{eq:3}
 U_\emph{}(x,y,z) =U_\emph{a}(x,y,z) +U_\emph{r}(x,y,z)
\end{equation}
\vspace{0.5em}

\vspace{-0.4em}
\begin{equation}
 \label{eq:4}
U_\emph{a}(x,y,z)=\xi \left \| P_\emph{current}-P_\emph{goal} \right \|^{2}
\end{equation}
\vspace{0.5em}
\vspace{-0.4em}
\begin{equation}
 \label{eq:5}
 U_\emph{r}(x,y,z) = \begin{cases}
 \frac{1}{2}\eta(\frac{1}{\rho(x,y,z)} - \frac{1}{d_\emph{0}})^2 \quad &, \hspace{2mm}\rho \leq d_\emph{0} \\
 0 \quad &,\hspace{2mm}\rho > d_\emph{0} \\
 
 \end{cases}
\end{equation}
where $U_\emph{}(x,y,z)$, $U_\emph{a}(x,y,z)$, and $U_\emph{r}(x,y,z)$ are the resulting, attraction, and obstacles' repulsive potential, respectively. The attraction potential is calculated as a function of current position $P_\emph{current}$ and the goal position $P_\emph{goal}$ where $\xi $ is the scaling factor. The obstacles' repulsive potential $U_\emph{r}(x,y,z)$ is only calculated when the drone is located within $d_\emph{0}$ distance (radius of influence) form other agents and is given by Eq. \ref{eq:5} where $\rho(x,y,z)$ is the chosen Euclidean distance function, and $\eta$ is the scaling factor. The repulsive potential is inversely proportional to the distance to other agents, therefore, the closer the drone the higher the repulsion until the drone is out of the radius of influence. The target position for each drone is calculated given the position of the leader drone and the required distance from the drone to the leader to maintain a formation for landing on the confined platform. In this work we opted for the Delta formation to fit the landing platform dimensions.

\subsection{Mobile Robot System Overview}

\begin{figure}[!h]
 \centering
 \includegraphics[width=0.8\linewidth]{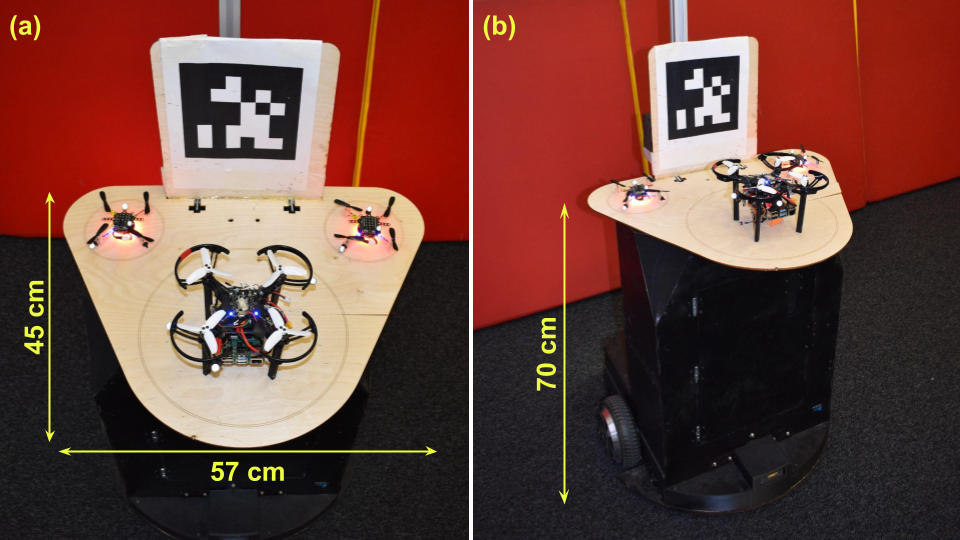} 
 \caption{(a) Landing pad for the leader (center) and follower (on both sides) drones. (b) Mobile robot represents a moving platform during the experiment.}
 \label{fig:mobile}
\end{figure}

Experiments were carried out on a differential drive mobile robot of cylindrical shape of 3 equipped with a drone landing pad (Fig. \ref{fig:mobile}). The robot landing pad with a capacity of three drones was located at the height of 70 cm and is equipped with a vertically placed visually distinctive tag. 

Intel NUC computer was used to run high-level algorithms in real-time. The robot was equipped with Hokuyo LIDAR to safely navigate in cluttered environment. BLDC-motor wheels are controlled through an STM-32f4 microcontroller.

\section{Experimental Setup}

A two-part experiment was carried out to evaluate the proposed approach of swarm landing: 1) landing on a static mobile robot and 2) landing on a moving mobile robot with several velocities. At the start of each experiment, the leader drone followed by other agents attempts to find the target. While in moving condition, the swarm tries to land on it. The swarm of drones used for the experiment includes one leader drone, which weighs 0.262 kg, including a 450 mAh 3S battery, and follower drones of 0.032 kg each including a 240 mAh 1S battery. Onboard processor of the leader drone is Crazyflie 2.1 firmware with a quad deck extension and Raspberry Pi 4 with a Pi camera recording RGB images of 640 x 480 pixels at a rate of 30 fps. Whereas the follower drones includes only the Crazyflie 2.1 controller. Each experiment was repeated 10 times resulting in overall 40 trials.

\subsection{Stationary Platform Experiment}

\subsubsection{Experiment} In this experiment, the mobile robot is set to a stationary position and the swarm of drones is launched. The leader drone has to find the April tag, and the follower drones moves accordingly to the received trajectory coordinates.
   
\subsubsection{Results} 
\begin{figure}[!h]
 \centering
 \includegraphics[width=0.8\linewidth]{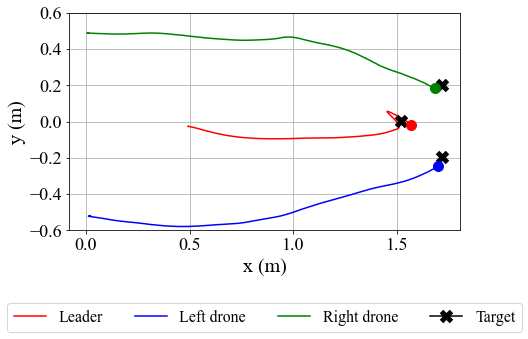} 
 \caption{Trajectories of drones landing on the stationary platform.}
 \label{fig:stationary}
\end{figure}

The drone landing trajectory is shown in Fig. \ref{fig:stationary}. After 10 landing experiments, the overall RMSE of swarm formation landing was 4.48 cm (Fig. \ref{fig:error}). The experimental results revealed that the drones landed within the close proximity to the target.
    
\subsection{Moving Platform Experiment}

\subsubsection{Experiment} 
\begin{figure}[!h]
 \centering
 \includegraphics[width=1.0\linewidth]{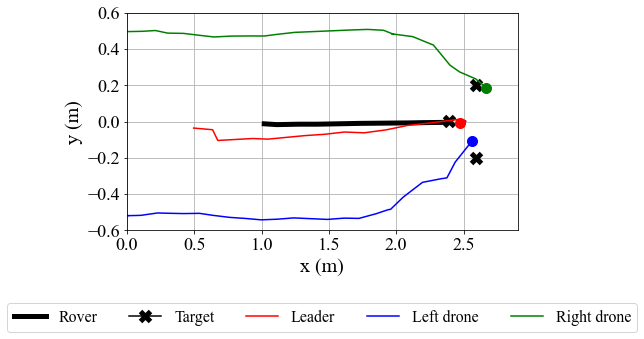} 
\caption{Trajectories of drones landing on a mobile platform with rover speed of 1.5 m/s.}
 \label{fig:mov_land}
\end{figure}

In this experiment, the mobile robot moves in a straight-line trajectory. This trajectory is calculated at different velocities of the mobile robot, i.e., 0.5 m/s, 1 m/s, and 1.5 m/s. As the mobile robot moves, the swarm of drones follows the robot's trajectory without changing their landing formation by yawing until they land on the target location. The flight trajectory taken for different courses and velocities is shown in Fig. \ref{fig:mov_land}
    
\subsubsection{Results} 
\begin{figure}[!h]
 \centering
 \includegraphics[width=0.7\linewidth]{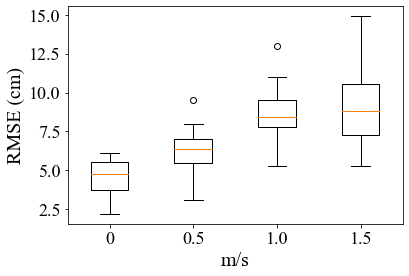} 
 \caption{RMSE of three drones landing on a mobile platform moving with different speed.}
 \label{fig:error}
\end{figure}

In this condition, we can observe that as the drones identify the target tag, they start to move toward it and go into the landing state as a certain threshold is achieved. The overall RMSE for the swarm formation at 0.5 m/s, 1.0 m/s and 1.5 m/s are 6.3 cm, 8.76 cm, and 8.98 cm, respectively (Fig. \ref{fig:error}). 10 experiments were conducted at different velocities of the mobile robot, for a total of 30 experiments. The experimental results revealed that drones land in close proximity to the target, however, the RMSE increases along with the speed of the moving platform.

\begin{table}[htbp]
\fontsize{10}{11}\selectfont
\caption{RMSE of Landing on Mobile Platform Moving with Three Different Velocities}

\centering
\begin{tabular}{|l|*{4}{c}}
\hline
 & \multicolumn{4}{c|}{Mobile robot velocity, m/s} \\ \cline{2-5} 
\makebox[30mm]{ } &\multicolumn{1}{c|}{0 }&\multicolumn{1}{c|}{0.5 }&\multicolumn{1}{c|}{1}&\multicolumn{1 }{c|}{1.5}\\
\hline

Leader drone RMSE, cm& \multicolumn{1}{c|}{3.97} & \multicolumn{1}{c|}{5.96} & \multicolumn{1}{c|}{8.21}& \multicolumn{1}{c|}{8.56} \\
\hline

Drone 2L RMSE, cm & \multicolumn{1}{c|}{4.83}& \multicolumn{1}{c|}{6.03} & \multicolumn{1}{c|}{9.27}& \multicolumn{1}{c|}{8.83} \\
\hline
Drone 3R RMSE, cm& \multicolumn{1}{c|}{4.53} &\multicolumn{1}{c|}{7.00} & \multicolumn{1}{c|}{8.32}&\multicolumn{1}{c|}{9.19} \\
\hline
\end{tabular}
\label{tab:Averaged experiment 2 results}
\end{table}

\subsubsection{Discussions} The experimental results revealed that the proposed CV-based ``leader-followers" strategy for the landing of a swarm on a moving platform is capable of safe landing for up to 1.5 m/s platform velocity, up to 4 m initial distance to the platform, and the RMSE threshold of 15 cm. Moreover, the RMSE does not depend on whether the drone performs as a leader or as a follower. Table \ref{tab:Averaged experiment 2 results} shows that the RMSE does not increase significantly when the rover velocity increases from 1.0 m/s to 1.5 m/s, as we limit the velocity of the drones to prevent the crash. Also, the velocity limit of the drones will explain the choice of the rover speed for the experiments. To support swarm landing on a mobile platform with higher velocity, it is required to investigate more complex target-estimation algorithms in a combination with APF and model-based state prediction approaches.

\section{Conclusions and Future Work}
We proposed a novel technology SwarmHive which allows a swarm of heterogeneous drones to land on a moving platform with a limited surface area. Artificial potential field approach ensured the the visual tag following and preventing the internal swarm collisions. The CV-based system was investigated to detect the visual tag and to calculate its distance from the drone. The experimental results revealed a low landing RMSE of 4.48 cm and 8.98 cm, in the case of static and moving platforms, respectively.

In the future work, we plan to study different flight conditions and external disturbances in our system that will allow extending the application of SwarmHive to outdoor environment. The technology can be applied for the long run exploration of the large-scale areas since the mobile robot can be equipped with a charging station.

The proposed heterogeneous swarm of drones can potentially significantly improve the scalability of the mission tasks, thus expanding the area of applications and flexibility of the functions that can be performed in real-time by accommodating data from different sensors and systems.

\bibliographystyle{IEEEtran}
\bibliography{sample}


\end{document}